\documentclass[letterpaper, 10 pt, conference]{ieeeconf}  % Comment this line out if you need a4paper

\IEEEoverridecommandlockouts                              % This command is only needed if 
\usepackage{times}
\usepackage{epsfig}
\usepackage{graphicx}
\usepackage{amsmath}
\usepackage{amssymb}
\usepackage{graphicx}
\usepackage{amsmath}
\usepackage{booktabs}
\usepackage{algorithm}
\usepackage{algorithmic}
\usepackage{comment}
\usepackage{bm} 

 \usepackage{multirow}

\title{\LARGE \bf
FlyCNS: Connectome-Grounded Information Organization for Communication-Constrained Embodied Control
}

\author{Jinchang Zhang, Jiakai Lin and Guoyu Lu
\thanks{Jinchang Zhang, Jiakai Lin and Guoyu Lu are with the Intelligent Vision and Sensing (IVS) Lab at Indiana University Bloomington.
{\tt\small guoyulu62@gmail.com}.}%
}
\begin{document}

\maketitle
\thispagestyle{empty}
\pagestyle{empty}

%%%%%%%%%%%%%%%%%%%%%%%%%%%%%%%%%%%%%%%%%%%%%%%%%%%%%%%%%%%%%%%%%%%%%%%%%%%%%%%%
\begin{abstract}
Robotic bodies are inherently distributed in sensing and actuation, yet learning-based control still commonly relies on centralized information processing. This work studies the problem of information organization in communication-constrained embodied control: which computations should remain local, and which information is worth transmitting for whole-body coordination.
We propose FlyCNS, an embodied information-organization framework inspired by the Drosophila brain--nerve-cord connectome. FlyCNS preserves local sensorimotor computation within each limb and enables selective long-range communication through separate ascending and descending routing pathways. From a real connectome, FlyCNS extracts the directional structural complexity of these two pathway types and uses it as a weak prior over communication allocation, while message content, transmission timing, and locomotion policies remain task-adaptive and are learned through reinforcement learning.
In Unitree Go1 simulation, FlyCNS exhibits more graceful performance degradation as the communication budget is tightened. Under the most restrictive setting, it uses only about 21--22\% of the communication of the full-communication reference, while still maintaining a tracking score of approximately 0.882 under both command protocols, with a gap of no more than 6.1\% from the full-communication reference. These results indicate that real neural connectomes can inform not only the structural design of control networks, but also provide transferable inductive biases for information organization across embodiments, guiding robots in balancing local computation and long-range coordination under limited communication resources.

\end{abstract}

\section{Introduction}
\label{sec:introduction}

A robot's body is inherently distributed, yet learning-based control policies often remain centralized in how information is processed.
In quadrupedal locomotion, for example, each leg must react rapidly to its own joint state and local feedback, while body-level motion objectives require coordination across limbs.
These processes jointly produce whole-body behavior, but they rely on different information and operate at different control scales.
A centralized policy exposes the complete body state to a single controller, implicitly assuming that all local information should continuously participate in global decision making.
Once computation is distributed across body modules, a more fundamental question arises:
which information must actually cross the boundary between local modules and a global coordinator to preserve effective whole-body coordination?

This question has a direct counterpart in distributed and modular robotic systems.
Leg-level or joint-level controllers can exchange state and control information with higher-level processors through shared communication buses such as CAN, RS-485~\cite{sprowitz2018oncilla}.
In such architectures, long-range information exchange is not free: bandwidth, latency, scheduling contention, and embedded resource constraints can all limit how frequently local state participates in global coordination.
We do not assume that the stock communication bus of the Unitree Go1 is itself saturated.
Instead, Go1 serves as a dynamical and morphological testbed, while an explicit communication budget models the limited information exchange that can arise between local computation and global coordination in distributed robots.
We therefore treat communication across the local--global boundary as a controlled information bottleneck rather than as a compression layer added after policy learning.

Existing modular and communication-aware policies have established important building blocks for this setting.
Local controllers can coordinate through learned messages, and communication can be adapted under bandwidth constraints~\cite{huang2020smp,guo2023demos,hu2020etcnet}.
However, under severe communication constraints, the policy must jointly discover how to control the body, what its messages should represent, and which information is worth transmitting across modules.
Learning all three solely from task reward turns the organization of information flow into an additional search problem coupled to control learning.
This motivates a further question:
can a task-independent structural prior provide a useful inductive bias for embodied information flow without prescribing the message content itself?

Biological nervous systems provide a natural instance shaped by evolution.
The Drosophila brain-and-nerve-cord connectome (BANC) reveals a multiscale sensorimotor organization in which high-rate, while ascending and descending pathways connect these local circuits to brain-level coordination~\cite{bates2026banc}.
This organization does not directly specify an optimal robot communication policy, but it provides a measurable structural hypothesis that is independent of the robot task.
Unlike connectome-based controllers that instantiate biological topology more directly~\cite{jin2026flygm,wang2026flynn}, we do not transfer individual neurons or specific biological circuits.
Instead, we extract the directional structural complexity of ascending and descending pathways from a predefined connectome subset and convert this statistic into a weak prior over local--global communication.
In other words, FlyCNS tests whether an information-organization principle can transfer across embodiments rather than whether a fly controller can be copied onto a robot.

\begin{figure*}[t]
\begin{center}
\includegraphics[width=17cm, height=7cm]{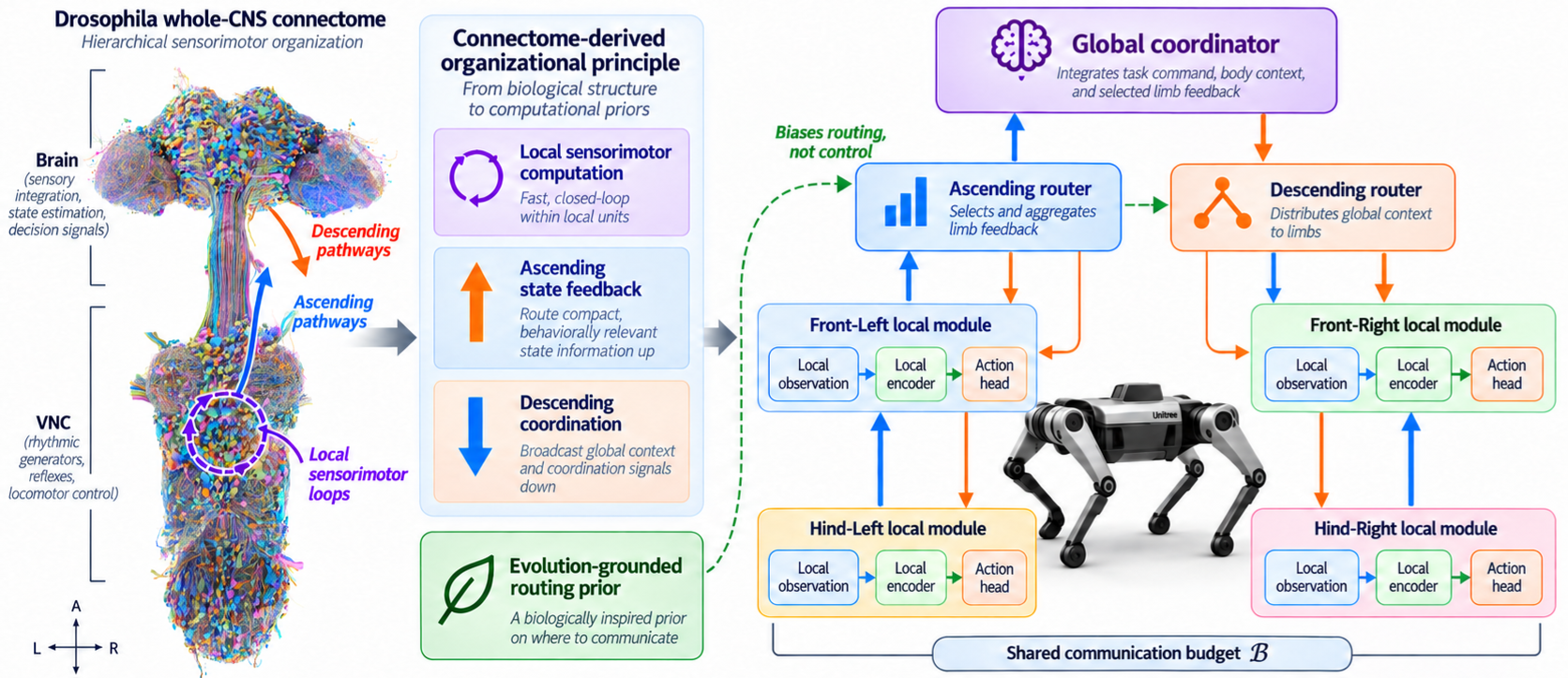}
\end{center}
\vspace{-6mm}
\caption{FlyCNS: from evolved neural organization to embodied information routing under constrained communication. The whole-CNS Drosophila connectome exhibits a hierarchical organization of information processing, in which ody-related sensorimotor processing is strongly organized within local circuits, while coordination across the body is mediated by ascending state feedback and descending modulation.
FlyCNS translates this principle into a local--global control architecture for quadruped robots and uses directional structural statistics extracted from the connectome as a weak routing prior to modulate ascending and descending information flow under a shared communication budget. Specific messages, routing decisions, and motor control remain task-adaptive and are learned through reinforcement learning. Whole-CNS rendering adapted from \cite{berg2026sexual}
}
\vspace{-5mm}
\label{archagent}
\end{figure*}

Based on this idea, we introduce \textbf{FlyCNS}, a connectome-informed learning framework for communication-constrained embodied control, illustrated in Fig.~\ref{archagent}.
FlyCNS preserves direct local sensorimotor computation within four limb modules while using a global coordinator for cross-limb information integration.
NeuroRoute separately learns selective transmission of ascending state feedback and descending coordination signals under a shared communication budget.
On top of this trainable local--global architecture, a fixed directional prior derived from BANC weakly biases the long-term allocation of the two information streams.
The message content, instantaneous routing decisions, and motor policy remain task-adaptive and are learned through robot--environment interaction.
Thus, biology constrains how information is organized rather than specifying which actions the robot should take.

We evaluate FlyCNS on simulated Unitree Go1 locomotion in MuJoCo Playground~\cite{zakka2025mujocoplayground}, explicitly measuring logical communication under three communication budgets.
At the most restrictive setting, FlyCNS uses only approximately 21--22\% of the communication required by the FULL reference while maintaining a tracking score of about $0.882$ under both Official and Scripted command protocols, remaining within $6.1\%$ of FULL.
At the 50\% and 25\% budgets, FlyCNS achieves both higher tracking performance and lower measured communication than a generic learned bidirectional router and a calibrated periodic-routing baseline.
As communication becomes increasingly scarce, FlyCNS degrades more gracefully, yielding a more favorable performance--communication trade-off.
Our main contributions are:

\begin{itemize}
    \item We formulate communication-constrained embodied control as a problem of allocating information between local sensorimotor computation and long-range coordination, making the local--global information boundary an explicit component of the control policy.

    \item We introduce \textbf{FlyCNS}, which transforms directional structure measured from a real Drosophila brain--cord connectome into a weak prior over learned ascending and descending communication, without requiring neuron-level or limb-level correspondence.

    \item We show that this organization yields a more favorable performance--communication trade-off for quadrupedal locomotion: FlyCNS maintains effective control under substantial communication reduction and exhibits more graceful performance degradation as available bandwidth is further reduced.
\end{itemize}

\section{Related Work}
\label{sec:related_work}

\subsection{Biological/Connectomic Control}
Biological control combines local sensorimotor processing with longer-range coordination.
The brain-and-nerve-cord (BANC) connectome makes this organization explicit through local sensory--effector pathways and ascending/descending connections~\cite{bates2026banc}.
Complementary brain reconstructions and cell-type annotations~\cite{dorkenwald2024wiring,schlegel2024annotation}, together with ventral-nerve-cord and premotor circuit analyses~\cite{azevedo2024vnc,lesser2024premotor}, provide the structural basis for studying how neural wiring supports movement.
Connectome-constrained models relate this wiring to sensorimotor and visual computation~\cite{shiu2024brain,lappalainen2024connectome}, while neuromechanical simulators connect neural controllers to realistic bodies and sensory feedback~\cite{lobato2022neuromechfly,wangchen2024neuromechfly,vaxenburg2025flybody}.
Biologically inspired recurrent controllers also support compact, interpretable autonomy and robust visual navigation~\cite{lechner2020ncp,chahine2023liquid}.
Moving toward direct use of measured topology, FlyGM~\cite{jin2026flygm} embeds a whole-brain connectomic graph in a policy for simulated fly locomotion, and FLYNN~\cite{wang2026flynn} structures a recurrent navigation policy with fly-brain connectivity.
These neuron-level models and topology-based policies leave open how to abstract brain--cord organization into a transferable routing prior for a robot with a different morphology.

\subsection{Modular/Decentralized Robot Learning}
Modular policies coordinate local controllers through shared parameters and information exchange.
Shared Modular Policies (SMP)~\cite{huang2020smp} combine reusable actuator policies with bottom-up and top-down messages, alongside graph-based approaches to morphology transfer and self-assembling control~\cite{wang2018nervenet,pathak2019assembling,whitman2023modular}.
However, morphology-constrained message passing is not uniformly beneficial, motivating learned alternatives to fixed physical connectivity~\cite{kurin2021cage}.
Insect-inspired decentralization has a separate lineage: reactive leg coordination~\cite{schilling2013walknet} precedes hexapod deep reinforcement learning with local and neighboring-leg observations~\cite{schilling2020hexapod}; related studies learn distributed controllers for articulated robots and examine the benefits of local information~\cite{sartoretti2019distributed,schilling2021local}.
DEMOS~\cite{guo2023demos} further reduces coupling by penalizing cross-branch motor contributions and removing weak connections while retaining essential ones.
Selective communication is also established in multi-agent learning, where event-triggered policies optimize transmission under bandwidth constraints~\cite{hu2020etcnet}.
Building on these modular and communication-efficient approaches, FlyCNS uses a connectome-derived directional prior to regularize the allocation of budgeted communication between limb controllers and a global coordinator.

\section{Method}
\label{sec:method}

FlyCNS transfers a deliberately coarse aspect of neural organization
from the Drosophila brain--cord system to robot control.
The transferred object is neither morphology nor biological circuitry,
but a directional prior over how long-range information is organized
between local sensorimotor computation and global coordination.
We instantiate this principle in three stages: limb-local modules
preserve direct sensorimotor computation, NeuroRoute learns when
information should cross the local--global boundary, and
connectome-derived statistics bias how scarce long-range communication
is distributed between ascending and descending traffic.
Message semantics, transmission timing, and motor behavior remain
task-adaptive and are learned for the robot.

\subsection{From Neural Organization to Embodied Information Architecture}
\label{sec:formulation}

The biological organization underlying FlyCNS suggests a division of
information processing into local sensorimotor computation, ascending
body-to-brain feedback, and descending brain-to-body coordination.
We transfer this organization at the level of information flow:
We transfer this organization at the level of information flow: local sensorimotor circuitry is abstracted as limb-local computation, ascending pathways as body-to-global information flow, and descending pathways as global-to-body coordination.
Consider a robot with physical state $s_t$, task command $u_t$, and
$N$ limb modules. Its actor observations are partitioned as
$
o_t =
\left(
o_t^G,
o_{1,t}^L,\ldots,o_{N,t}^L
\right),
$
where $o_{i,t}^L$ contains the proprioceptive information locally
available to limb $i$, while $o_t^G$ contains body-level state and the
task command available to the global coordinator. In Go1, $N=4$ and
each limb controls three actuated joints.
This partition creates an explicit information boundary.
Local observations remain directly available to local control, whereas
limb-specific information can affect global coordination only through
the ascending interface. Conversely, global task context reaches local
action generation through descending communication. Communication
therefore determines which remote information becomes available to the
policy.
Let $r_t$ denote the task reward and $C_t$ the logical communication
crossing this boundary. 
\begin{equation}
\small
\max_{\theta}\quad
J(\theta)
=
\mathbb{E}_{\pi_\theta}
\left[
\sum_{t=0}^{T-1}\gamma^t r_t
\right],
\text{s.t.}\quad
\bar C(\theta)
=
\frac{1}{T}
\mathbb{E}_{\pi_\theta}
\left[
\sum_{t=0}^{T-1} C_t
\right]
\le B ,
\label{eq:problem}
\end{equation}
where $B$ is the average communication budget.
The control problem therefore concerns action generation and information allocation: computation can remain local, while only selected information is exposed to whole-body coordination.

\subsection{Local Sensorimotor Autonomy and Global Coordination}
\label{sec:architecture}

Each limb first encodes its local observation together with a limb
identity embedding,
$
h_{i,t}^L
=
E\left([o_{i,t}^L,e_i]\right),
\label{eq:local_encoding}
$
where $e_i$ denotes the leg identity. The local encoder is shared
across limbs. From this representation, the limb forms a candidate
ascending message,
$
m_{i,t}^{\uparrow}
=M^{\uparrow}(h_{i,t}^L).
$
After the ascending communication state is updated, the global
coordinator combines body-level observations with the currently
available limb messages,
\begin{equation}
h_t^G
=
F\left(
o_t^G,
\left\{
\bar m_{i,t}^{\uparrow},
\delta_{i,t}^{\uparrow}
\right\}_{i=1}^{N}
\right),
\label{eq:global_representation}
\end{equation}
where $\bar m_{i,t}^{\uparrow}$ denotes the receiver-side cached
ascending message and $\delta_{i,t}^{\uparrow}$ summarizes its age.
The coordinator then generates limb-specific descending messages,
\begin{equation}
\left[
m_{1,t}^{\downarrow},\ldots,m_{N,t}^{\downarrow}
\right]
=
M^{\downarrow}(h_t^G).
\label{eq:descending_message}
\end{equation}

Each motor head produces action of one limb from its current local
representation and  available descending context,
\begin{equation}
a_{i,t}
\sim
\pi_A
\left(
\cdot
\mid
h_{i,t}^L,
\bar m_{i,t}^{\downarrow},
\delta_{i,t}^{\downarrow}
\right),
a_t=
\operatorname{concat}_{i=1}^{N} a_{i,t}.
\label{eq:local_action}
\end{equation}

The resulting information path can be summarized as
\begin{equation}
o_{i,t}^{L}
\rightarrow
h_{i,t}^{L}
\rightarrow
m_{i,t}^{\uparrow}
\rightarrow
h_t^G
\rightarrow
m_{i,t}^{\downarrow}
\rightarrow
a_{i,t}.
\label{eq:information_chain}
\end{equation}
The two message transitions are the points at which information crosses
the local--global boundary. Global coordination therefore augments
local sensorimotor computation rather than replacing it.

\subsection{NeuroRoute: Selective Bidirectional Coordination}
\label{sec:routing}

NeuroRoute determines when the two long-range information streams
should be updated. Ascending and descending communication occupy
different positions in the control loop: ascending messages update what
the coordinator knows about the body, whereas descending messages
update the global context available to local action generation.
We therefore parameterize the two routing decisions independently.
The ascending router operates directly on the local representation,
$
\xi_{i,t}^{\uparrow}=h_{i,t}^L,
\ell_{i,t}^{\uparrow}
=
R^{\uparrow}(\xi_{i,t}^{\uparrow}),
$
while the descending router acts on the global coordination state
together with the target limb identity,
$
\xi_{i,t}^{\downarrow}
=
[h_t^G,e_i],
\ell_{i,t}^{\downarrow}
=
R^{\downarrow}(\xi_{i,t}^{\downarrow}).
$
At the target operating budget, the corresponding transmission
probabilities are
$
p_{i,t}^{d}
=
\sigma(\ell_{i,t}^{d}),
d\in\{\uparrow,\downarrow\},
$
and hard routing decisions are sampled as
$
g_{i,t}^{d}
\sim
\operatorname{Bernoulli}(p_{i,t}^{d}).
$
Ascending gates are sampled first; the resulting ascending cache state
is then used to compute the global representation and the conditional
descending routing probabilities. Thus, routing follows the causal
order of the control loop rather than treating all communication
decisions as independent.
When a message is not transmitted, the receiver retains its most
recent value,
\begin{equation}
\bar m_{i,t}^{d}
=
g_{i,t}^{d}m_{i,t}^{d}
+
\left(1-g_{i,t}^{d}\right)\bar m_{i,t-1}^{d}.
\label{eq:message_cache}
\end{equation}
Its age is provided as lightweight temporal context so that recently
updated information can be distinguished from stale information.
A transmitted packet contains 32 float32 values and four logical
metadata bits, yielding
\begin{equation}
C_t^{d}
=
1028\sum_{i=1}^{N}g_{i,t}^{d},
\qquad
C_t=
C_t^{\uparrow}+C_t^{\downarrow}.
\label{eq:communication_cost}
\end{equation}
With four ascending and four descending channels,
full communication corresponds to
$C_{\mathrm{full}}=8{,}224$ logical bits per control step.
We optimize the average communication constraint using a standard
Lagrangian actor--critic formulation.
During training, the minimum transmission probability is progressively
relaxed from full communication toward the learned routing policy as
the communication budget is tightened.
This curriculum stabilizes learning under severe sparsity without
changing the final routing objective.

\subsection{Connectome-Grounded Directional Prior}
\label{sec:bio_prior}

The preceding architecture can learn its information organization
entirely from robot interaction. FlyCNS instead introduces a
task-independent directional preference extracted from the
Drosophila brain--cord connectome before robot optimization.
We construct two independently measured directional connectivity
representations:
$
W^{\uparrow}\in\mathbb{R}_{\ge0}^{3\times16},
W^{\downarrow}\in\mathbb{R}_{\ge0}^{16\times3},
$
where $W^{\uparrow}$ summarizes connectivity from three
leg-related sensory groups to 16 ascending super-clusters, and
$W^{\downarrow}$ summarizes connectivity from the corresponding
16 descending super-clusters to three leg-related motor groups.
The two matrices are extracted and normalized using the same frozen
type-balanced protocol before robot training.
To obtain a scale-normalized description of how broadly structural
influence is distributed in each direction, we compute the normalized
singular spectrum
\begin{equation}
q_j^d
=
\frac{\sigma_j(W^d)}
{\sum_l \sigma_l(W^d)}
\label{eq:normalized_spectrum}
\end{equation}
and its effective rank
\begin{equation}
r_d
=
\exp
\left(
-\sum_j q_j^d\log q_j^d
\right),
\qquad
d\in\{\uparrow,\downarrow\}.
\label{eq:effective_rank}
\end{equation}

The two structural complexities define the directional prior
\begin{equation}
p_{\mathrm{bio}}^d
=
\frac{r_d}
{r_{\uparrow}+r_{\downarrow}},
p_{\mathrm{bio}}
=
(0.5544605,\;0.4455395)
\end{equation}
for ascending and descending communication, respectively.
FlyCNS transfers this statistic at the same coarse level: it biases the
relative allocation of long-range communication, rather than fixing
individual transmissions.
For each state, let the expected ascending send rate be
\begin{equation}
u_{\theta,t}
=
\frac{1}{N}
\sum_{i=1}^{N}
p_{i,t}^{\uparrow},
\label{eq:up_rate}
\end{equation}
and let the expected descending send rate marginalize over the possible
ascending routing patterns,
\begin{equation}
d_{\theta,t}
=
\mathbb{E}_{g_t^{\uparrow}}
\left[
\frac{1}{N}
\sum_{i=1}^{N}
p_{i,t}^{\downarrow}
\left(g_t^{\uparrow}\right)
\right].
\label{eq:down_rate}
\end{equation}

Using minibatch averages $\bar u_\theta$ and $\bar d_\theta$, we define
the learned directional allocation
\begin{equation}
\rho_\theta
=
\frac{
(\bar u_\theta+\epsilon,\,
 \bar d_\theta+\epsilon)
}{
\bar u_\theta+\bar d_\theta+2\epsilon
},
\qquad
\epsilon=10^{-8}.
\label{eq:robot_allocation}
\end{equation}

The biological prior is introduced through
\begin{equation}
\mathcal R_{\mathrm{bio}}(\theta)
=
D_{\mathrm{KL}}
\left(
\rho_\theta
\;\Vert\;
p_{\mathrm{bio}}
\right),
\label{eq:bio_regularizer}
\end{equation}
with a fixed coefficient $\beta=0.01$.
The resulting training objective can be written as
\begin{equation}
\mathcal L
=
\mathcal L_{\mathrm{PPO}}
+
\lambda
\left(
\bar C-B
\right)
+
0.01\,
D_{\mathrm{KL}}
\left(
\rho_\theta
\Vert
p_{\mathrm{bio}}
\right),
\label{eq:flycns_objective}
\end{equation}
where the communication term controls the total amount of long-range
traffic and the biological term biases its directional organization.
The two terms therefore act at different levels:
the communication budget determines how much information can cross the
local--global boundary, whereas the connectome prior biases how this
limited communication is distributed between ascending and descending
coordination.
The content and timing of individual messages remain state-dependent
and task-adaptive.
In FlyCNS, evolution supplies a prior over information organization;
robot interaction supplies its task-specific semantics and control.

\begin{table*}[t]
\centering
\setlength{\tabcolsep}{6pt}
\resizebox{0.94\textwidth}{!}{
\begin{tabular}{clccrrcc}
\toprule
\multirow{2}{*}{Nominal budget} & \multirow{2}{*}{Method} & \multicolumn{2}{c}{Tracking $S\uparrow$} & \multicolumn{2}{c}{Communication $C\downarrow$} & \multicolumn{2}{c}{Pass / 3} \\
\cmidrule(lr){3-4}\cmidrule(lr){5-6}\cmidrule(lr){7-8}
 & & O & S & O & S & O & S \\
\midrule
Reference & Central & $0.933\pm0.002$ & $0.936\pm0.002$ & -- & -- & 3 & 3 \\
 & FULL & $0.934\pm0.002$ & $0.939\pm0.001$ & 8,224 & 8,224 & 3 & 3 \\
\midrule
75\% & ASYM & $0.908\pm0.001$ & $0.917\pm0.002$ & 6,170 & 6,169 & 3 & 3 \\
 & PERIODIC & $0.916\pm0.001$ & $0.925\pm0.003$ & 6,169 & 6,169 & 3 & 3 \\
 & \textbf{FlyCNS} & $\mathbf{0.926\pm0.003}$ & $\mathbf{0.930\pm0.004}$ & 5,480 & 5,278 & 3 & 3 \\
\midrule
50\% & ASYM & $0.700\pm0.064$ & $0.875\pm0.004$ & 4,144 & 4,143 & 0 & 3 \\
 & PERIODIC & $0.819\pm0.033$ & $0.892\pm0.001$ & 4,143 & 4,143 & 2 & 3 \\
 & \textbf{FlyCNS} & $\mathbf{0.918\pm0.001}$ & $\mathbf{0.919\pm0.003}$ & 3,870 & 3,659 & 3 & 3 \\
\midrule
25\% & ASYM & $0.436\pm0.006$ & $0.767\pm0.002$ & 2,091 & 2,092 & 0 & 0 \\
 & PERIODIC & $0.547\pm0.077$ & $0.840\pm0.018$ & 2,090 & 2,090 & 0 & 1 \\
 & \textbf{FlyCNS} & $\mathbf{0.882\pm0.004}$ & $\mathbf{0.882\pm0.006}$ & 1,823 & 1,755 & 3 & 3 \\
\bottomrule
\end{tabular}
}
\caption{Final control--communication results. O/S denote official/scripted commands; $S$ is the tracking score and $C$ is realized evaluation communication in logical bits per control step. Scores are mean $\pm$ sample SD across three development training seeds. ``Pass'' counts seeds meeting the numerical locomotion criteria in each mode, not episode survival. }
\vspace{-5mm}
\label{tab:main}
\end{table*}

\begin{figure*}[t]
\centering
\begin{minipage}[t]{0.49\textwidth}
\centering
\includegraphics[width=\linewidth]{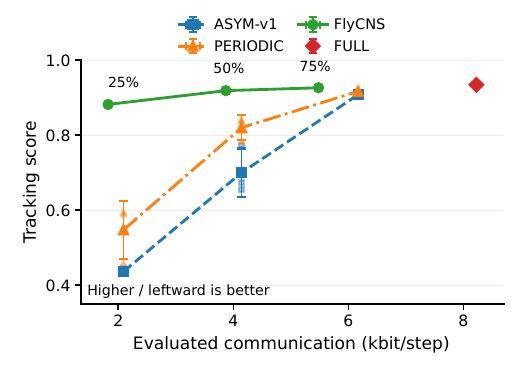}
\vspace{-6mm}
\par\small (a) Official commands
\end{minipage}\hfill
\begin{minipage}[t]{0.49\textwidth}
\centering
\includegraphics[width=\linewidth]{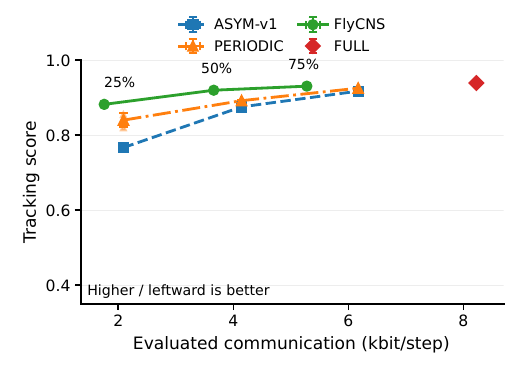}
\vspace{-6mm}
\par\small (b) Scripted commands
\end{minipage}
\caption{Observed control--communication trade-offs. FULL is an all-send reference configuration.}
\vspace{-5mm}
\label{fig:pareto}
\end{figure*}

\section{Experiments}
\subsection{Experimental Protocol}
\label{sec:setup}

\textbf{Task and training.}
We use the Unitree Go1 robot in MuJoCo Playground's {Go1JoystickFlatTerrain} task~\cite{zakka2025mujocoplayground}. The policy controls 12 joint targets and is optimized using a PPO-based learning framework~\cite{schulman2017ppo}, with a privileged critic and running observation normalization. Each training run consists of \(206{,}438{,}400\) environment interactions.

\textbf{Comparisons.}
Central uses a centralized actor. FULL consists of four limb-local modules and a global coordinator, transmitting all ascending and descending messages at every control step. ASYM uses learned bidirectional communication routing and serves as a generic adaptive-routing baseline. PERIODIC follows a fixed periodic communication schedule whose transmission frequency is calibrated from ASYM  communication records, without selecting the communication period based on task performance.

\textbf{Evaluation.}
Each checkpoint is evaluated using eight fixed resets under each command mode, with each rollout lasting \(1{,}000\) control steps, corresponding to \(20\,\mathrm{s}\). Official commands are generated by the environment's default command generator. Scripted commands switch every \(5\,\mathrm{s}\) among forward motion
\((v_x^*=0.8\,\mathrm{m/s})\), lateral motion
\((v_y^*=0.4\,\mathrm{m/s})\), in-place turning
\((\omega_z^*=0.8\,\mathrm{rad/s})\), and combined forward motion with reverse turning
\((v_x^*=0.6\,\mathrm{m/s},\,\omega_z^*=-0.6\,\mathrm{rad/s})\); all remaining command components are set to zero. Metrics are first averaged across resets and then across training seeds. Error bars report the sample standard deviation across three training seeds.

\begin{table*}[t]
\centering
\setlength{\tabcolsep}{5.3pt}
\resizebox{0.9\textwidth}{!}{
\begin{tabular}{clcrrrrrr}
\toprule
Budget & Method & Mode &
$R_v\uparrow$ &
NS (\%)$\downarrow$ &
$e_v\downarrow$ &
$e_\omega\downarrow$ &
$C_\uparrow\downarrow$ &
$C_\downarrow\downarrow$ \\
\midrule

75\%
& ASYM & O & 0.915 & 0.072 & 0.162 & 0.109 & 3,077 & 3,093 \\
&          & S & 0.909 & 0.100 & 0.136 & 0.104 & 3,077 & 3,093 \\
& PERIODIC & O & 0.914 & 0.055 & 0.157 & 0.102 & 3,077 & 3,093 \\
&          & S & 0.906 & 0.217 & 0.130 & 0.094 & 3,077 & 3,093 \\
& \textbf{FlyCNS}
           & O & \textbf{0.930} & \textbf{0.047} & \textbf{0.145} & 0.096 & \textbf{2,548} & \textbf{2,932} \\
&          & S & \textbf{0.923} & 0.183 & \textbf{0.125} & \textbf{0.092} & \textbf{2,465} & \textbf{2,813} \\
\midrule

50\%
& ASYM & O & 0.694 & 10.036 & 0.474 & 0.150 & 2,056 & 2,089 \\
&          & S & 0.834 & 0.283  & 0.154 & 0.143 & 2,054 & 2,089 \\
& PERIODIC & O & 0.819 & 4.515  & 0.304 & 0.142 & 2,055 & 2,089 \\
&          & S & 0.858 & 0.178  & 0.147 & 0.123 & 2,055 & 2,089 \\
& \textbf{FlyCNS}
           & O & \textbf{0.927} & \textbf{0.049} & \textbf{0.150} & \textbf{0.103} & \textbf{1,724} & 2,146 \\
&          & S & \textbf{0.920} & \textbf{0.161} & \textbf{0.128} & \textbf{0.104} & \textbf{1,609} & \textbf{2,050} \\
\midrule

25\%
& ASYM & O & 0.371 & 30.076 & 0.691 & 0.178 & 1,037 & 1,054 \\
&          & S & 0.652 & 8.950  & 0.236 & 0.176 & 1,037 & 1,055 \\
& PERIODIC & O & 0.503 & 20.286 & 0.633 & 0.158 & 1,037 & 1,054 \\
&          & S & 0.771 & 0.717  & 0.186 & 0.151 & 1,037 & 1,054 \\
& \textbf{FlyCNS}
           & O & \textbf{0.916} & \textbf{0.019} & \textbf{0.166} & \textbf{0.132} & \textbf{803} & \textbf{1,020} \\
&          & S & \textbf{0.912} & \textbf{0.128} & \textbf{0.142} & \textbf{0.137} & \textbf{743} & \textbf{1,012} \\

\bottomrule
\end{tabular}
}
\caption{
Motion quality and directional communication statistics across communication budgets.
Entries are means over three training seeds; O/S denote official/scripted commands.
$R_v$ is realized/commanded planar speed, NS is the near-static percentage,
and $e_v$ and $e_\omega$ are RMSE in m/s and rad/s.
$C_\uparrow$ and $C_\downarrow$ denote measured ascending and descending
logical communication in bits per control step.
}
\vspace{-7mm}
\label{tab:additional}
\end{table*}

\textbf{Metrics.}
In addition to the environment task reward, we use a fixed-horizon tracking score \(S\) to evaluate motion-tracking performance:
\begin{equation}
S=\frac{1}{T}\sum_{t=1}^{T} w_t
\exp\!\left[-\frac{\|\mathbf v_t-\mathbf v_t^*\|_2^2}{\sigma_v^2}
-\frac{(\omega_{z,t}-\omega_{z,t}^*)^2}{\sigma_\omega^2}\right],
\label{eq:score}
\end{equation}
where
$
\sigma_v=0.5\,\mathrm{m/s},
\sigma_\omega=0.5\,\mathrm{rad/s},
$
and \(w_t\) is set to zero from the first failure onward. Thus, \(S\) measures motion-tracking quality over a fixed evaluation horizon rather than binary task success.
Motion feasibility is further evaluated using the realized-to-commanded speed ratio \(R_v\), near-static fraction, planar-velocity RMSE, yaw-rate RMSE, and episode completion. \(R_v\) and the near-static fraction are computed only over surviving steps with commanded planar speed of at least \(0.2\,\mathrm{m/s}\), where near-static behavior is defined as an actual speed below \(0.05\,\mathrm{m/s}\). Pure turning commands remain included in \(S\) and yaw-rate RMSE.

\textbf{Communication.}
Each transmitted message contains 32 float32 values and 4 metadata bits, corresponding to
1028 bits
With four limbs and two communication directions, FFULL transmits 8,224 logical bits/control step.
Communication cost is measured from the packets actually transmitted, including message headers; reusing a cached message incurs no additional communication cost. Central has no corresponding cost under this message interface.
The nominal 75\%, 50\%, and 25\% budgets indicate training configurations rather than measured communication rates. Communication efficiency is therefore compared using the empirically observed performance--communication trade-off.

\subsection{Performance--Communication Trade-off}
\label{sec:tradeoff}

The experimental results are summarized in Table~\ref{tab:main} and Fig.~\ref{fig:pareto}. FULL achieves tracking performance comparable to Central: under Official / Scripted commands, FULL obtains scores of 0.9339 / 0.9386, compared with 0.9327 / 0.9362 for Central. These results demonstrate that the proposed local--global architecture can learn effective locomotion policies.
Compared with the two baselines under the same communication budgets, FlyCNS timproves mean tracking performance while reducing realized communication. Relative to PERIODIC, under Official commands, the tracking-score improvement increases from 0.0095 at the 75\% budget to 0.0988 at 50\% and 0.3344 at 25\%, while communication is reduced by 11.17\%, 6.59\%, and 12.81\%, respectively. Under Scripted commands, the corresponding tracking improvements are 0.0054, 0.0278, and 0.0421, accompanied by communication reductions of 14.46\%, 11.69\%, and 16.04\%, respectively.
At the 50\% and 25\% budgets, FlyCNS tachieves both higher tracking performance and lower communication cost than either baseline across all seed--mode comparisons. At the 75\% budget, in one Scripted comparison with PERIODIC, FlyCNS thas a 0.000123 lower tracking score while still using less communication.

\begin{figure*}[t]
\centering
\includegraphics[width=16cm, height=10cm]{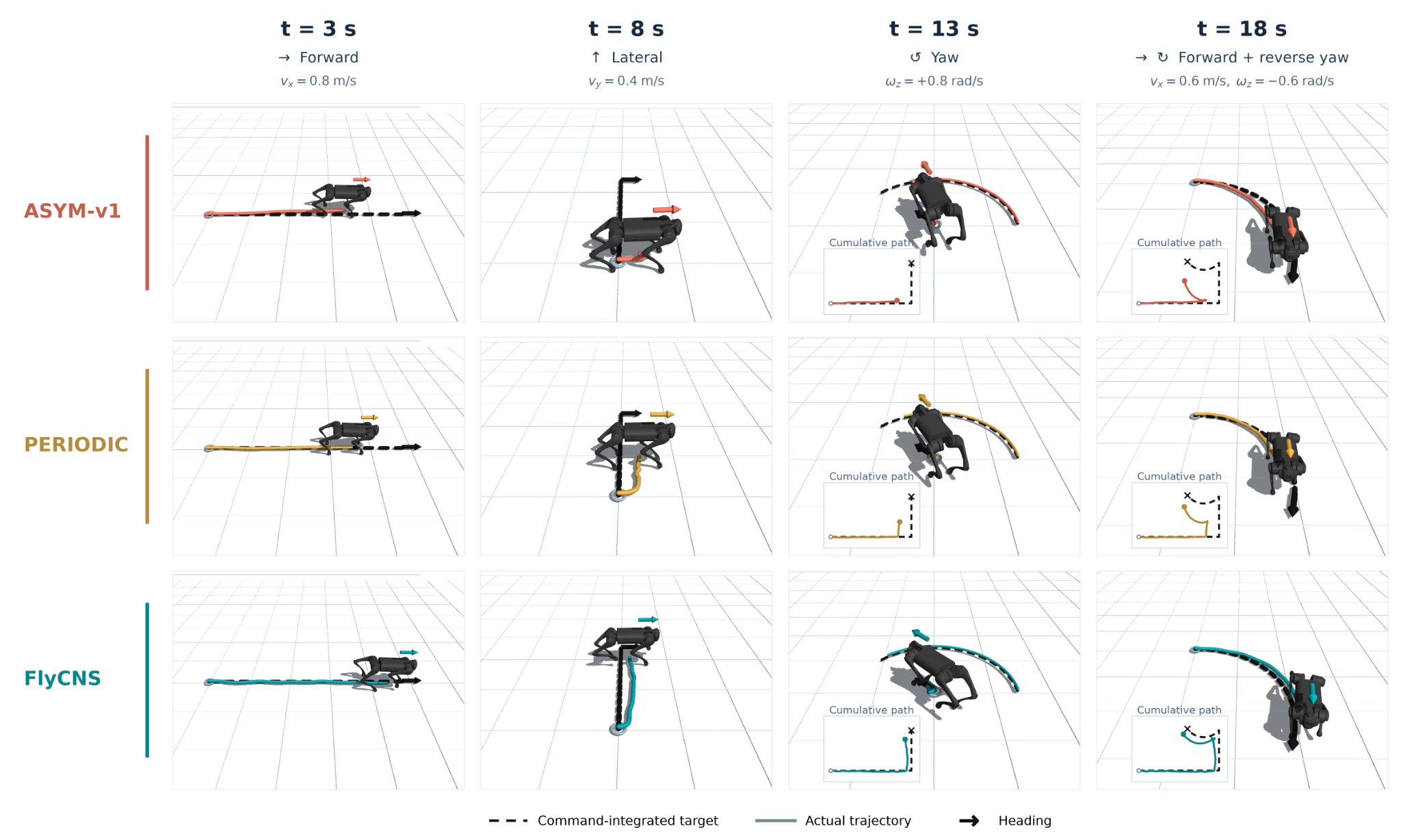}
\vspace{-5mm}
\caption{Comparison under the 25\% nominal budget. Columns correspond to the forward, lateral, yaw, and combined forward/reverse-yaw command phases at $t=3,8,13,18\,\mathrm{s}$. Dashed black curves indicate command-integrated reference trajectories, colored curves show the actual trajectories, and arrows indicate robot heading. Insets show the cumulative motion paths.}
\vspace{-4mm}
\label{fig:qualitative}
\end{figure*}

\begin{figure*}[t]
\begin{center}
%\framebox[4.0in]{$\;$}
\includegraphics[width=16cm, height=6cm]{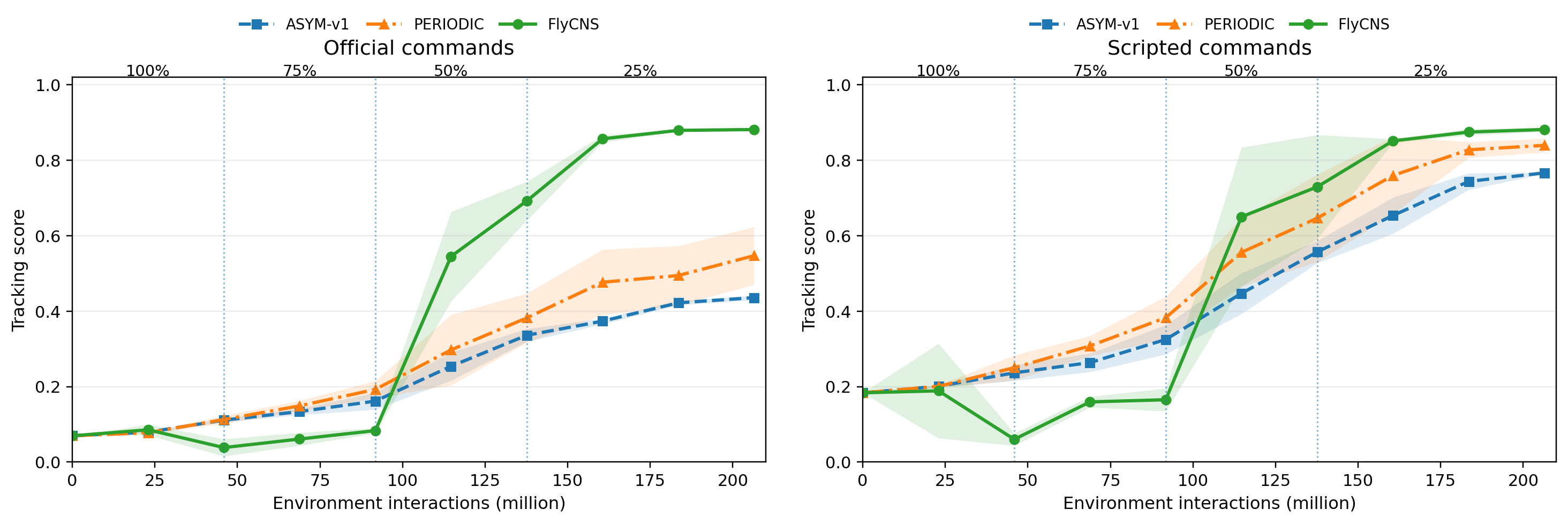}
\end{center}
\vspace{-6 mm}
\caption{Learning dynamics for configurations targeting 25\% communication. Top labels indicate the FlyCNS training curriculum; every policy is evaluated at the final 25\% communication target.}
\label{fig:learning25}
\vspace{-5mm}
\end{figure*}

\subsection{Control Quality under Increasing Communication Scarcity}
\label{sec}

Table~\ref{tab:additional} shows that the differences in control quality become increasingly pronounced as the communication budget is reduced from 75\% to 25\%. At the 75\% budget, all three methods maintain relatively strong locomotion performance, while FlyCNS already achieves lower realized communication. When the budget is reduced to 50\%, ASYM and PERIODIC exhibit noticeable degradation, particularly under Official commands, as reflected by lower speed ratios and higher near-static fractions. In contrast, FlyCNS still maintains $R_v$=$0.927/0.920$, with near-static fractions below 0.2\%.
The separation becomes most pronounced at the 25\% budget. FlyCNS achieves $R_v=0.916/0.912$ under Official / Scripted commands, with near-static fractions of only 0.019\% / 0.128\%. Its realized communication is further reduced to 1,823 / 1,755 bits per control step, corresponding to only 22.16\% / 21.34\% of FULL. Moreover, all three FlyCNS training seeds satisfy the complete numerical locomotion criteria under both command protocols, whereas ASYM fails in both modes and PERIODIC passes only for one Scripted-command seed. These results indicate that, under severe communication scarcity, the baselines exhibit substantial motion stagnation and command-tracking degradation, while FlyCNS remains in a feasible locomotion regime.

Fig.~\ref{fig:qualitative} provides a complementary view through synchronized rollouts. Although the methods can exhibit similarly upright postures at individual time steps, their cumulative trajectories diverge substantially. In particular, during the lateral, yaw, and combined-motion phases, the trajectories of  ASYM and PERIODIC increasingly deviate from the command-integrated reference, whereas FlyCNS tremains closer to the desired motion. Thus, low-bandwidth failure does not necessarily appear as an immediate fall; more often, the robot remains upright and moving while no longer faithfully executing the high-level command.

Communication compression also incurs a performance cost for FlyCNS itself. Reducing the budget from 50\% to 25\% decreases realized communication by 52.91\% / 52.04\% under Official / Scripted commands, while the tracking score drops by only 0.0367 / 0.0377. FlyCNS therefore does not eliminate the fundamental trade-off between communication and control performance; rather, it degrades more gracefully as communication becomes scarce, yielding a more favorable performance--communication trade-off.

\subsection{Learning Dynamics, Traffic, and Attribution} \label{sec:dynamics} Figure~\ref{fig:learning25} shows the performance evolution of policies at different stages of training under a unified final 25\% communication budget. FlyCNS does not lead during the early stage: its Official tracking score is only 0.0834 at 91.75M interactions, but then increases rapidly to 0.5452 at 114.69M and 0.8569 at 160.56M, finally reaching 0.8815, clearly higher than 0.5471 for PERIODIC and 0.4359 for  ASYM. A similar late-stage improvement is also observed under Scripted commands.

\section{Conclusion}
\label{sec:conclusion}

This work studies communication-constrained embodied control as an
information-organization problem: which computations should remain
local, and which information is worth transmitting for whole-body
coordination. We introduced \textbf{FlyCNS}, which combines limb-local
sensorimotor computation with selective ascending and descending
communication, and transfers a directional structural prior extracted
from the Drosophila brain--cord connectome to robot routing.
Rather than copying biological circuitry, FlyCNS uses the connectome
to bias how limited long-range communication is organized, while
message semantics, transmission timing, and motor behavior remain
learned from robot interaction.
Experiments on Unitree Go1 locomotion show that FlyCNS consistently
improves the performance--communication trade-off as communication
becomes scarce. At the most restrictive setting, the controller
retains effective locomotion using only about 21--22\% of the
full-communication traffic, while exhibiting substantially more
graceful degradation than generic learned routing and periodic
scheduling. These results suggest
that biological connectomes can contribute more than architectural
inspiration: measured principles of neural information organization
can serve as transferable inductive biases for how robots distribute
information across their bodies.

%%%%%%%%%%%%%%%%%%%%%%%%%%%%%%%%%%%%%%%%%%%%%%%%%%%%%%%%%%%%%%%%%%%%%%%%%%%%%%%%

\bibliographystyle{IEEEtran}
\bibliography{egbib}

\end{document}